\documentclass[sigconf]{acmart}

\AtBeginDocument{%
	\providecommand\BibTeX{{%
			\normalfont B\kern-0.5em{\scshape i\kern-0.25em b}\kern-0.8em\TeX}}}

\copyrightyear{2022} 
\acmYear{2022} 
\setcopyright{acmcopyright}\acmConference[WWW '22]{Proceedings of the ACM Web Conference 2022}{April 25--29, 2022}{Virtual Event, Lyon, France}
\acmBooktitle{Proceedings of the ACM Web Conference 2022 (WWW '22), April 25--29, 2022, Virtual Event, Lyon, France}
\acmPrice{15.00}
\acmDOI{10.1145/3485447.3511936}
\acmISBN{978-1-4503-9096-5/22/04}

\usepackage{graphicx}
\usepackage{enumitem}

\begin{document}
	
\title{Conditional Generation Net for Medication Recommendation}

\author{Rui Wu}
\authornote{This work was done when Rui Wu was an intern at Tencent.}
\email{rhyswu@outlook.com}
\affiliation{%
  \institution{Southeast University}
  \country{China}
}

\author{Zhaopeng Qiu}
\email{zhaopengqiu@tencent.com}
\affiliation{%
  \institution{Tencent}
  \country{China}
}

\author{Jiacheng Jiang}
\email{acejcjiang@tencent.com}
\affiliation{%
  \institution{Tencent}
  \country{China}
}

\author{Guilin Qi}
\email{gqi@seu.edu.cn}
\affiliation{%
  \institution{Southeast University}
  \country{China}
}

\author{Xian Wu}
\authornote{Xian Wu is the Corresponding Author.}
\email{kevinxwu@tencent.com}
\affiliation{%
  \institution{Tencent}
  \country{China}
}
	
\renewcommand{\shortauthors}{Wu and Qiu, et al.}

\begin{abstract}
Medication recommendation targets to provide a proper set of medicines according to patients' diagnoses, which is a critical task in clinics. Currently, the recommendation is manually conducted by doctors. However, for complicated cases, like patients with multiple diseases at the same time, it's difficult to propose a considerate recommendation even for experienced doctors. This urges the emergence of automatic medication recommendation which can help treat the diagnosed diseases without causing harmful drug-drug interactions. 
Due to the clinical value, medication recommendation has attracted growing research interests. 
Existing works mainly formulate medication recommendation as a multi-label classification task to predict the set of medicines. In this paper, we propose the Conditional Generation Net (COGNet) which introduces a novel {\em copy-or-predict} mechanism to generate the set of medicines. 
Given a patient, the proposed model first retrieves his or her historical diagnoses and medication recommendations and mines their relationship with current diagnoses. Then in predicting each medicine, the proposed model decides whether to copy a medicine from previous recommendations or to predict a new one. This process is quite similar to the decision process of human doctors. We validate the proposed model on the public MIMIC data set, and the experimental results show that the proposed model can outperform state-of-the-art approaches.
\end{abstract}

\ccsdesc[500]{Information systems~Data mining}
\ccsdesc[500]{Applied computing~Health informatics}

\keywords{medication recommendation, electronic health record, generation}

\maketitle

\section{Introduction}
\begin{figure}[h]
	\centering
	\includegraphics[width=\linewidth]{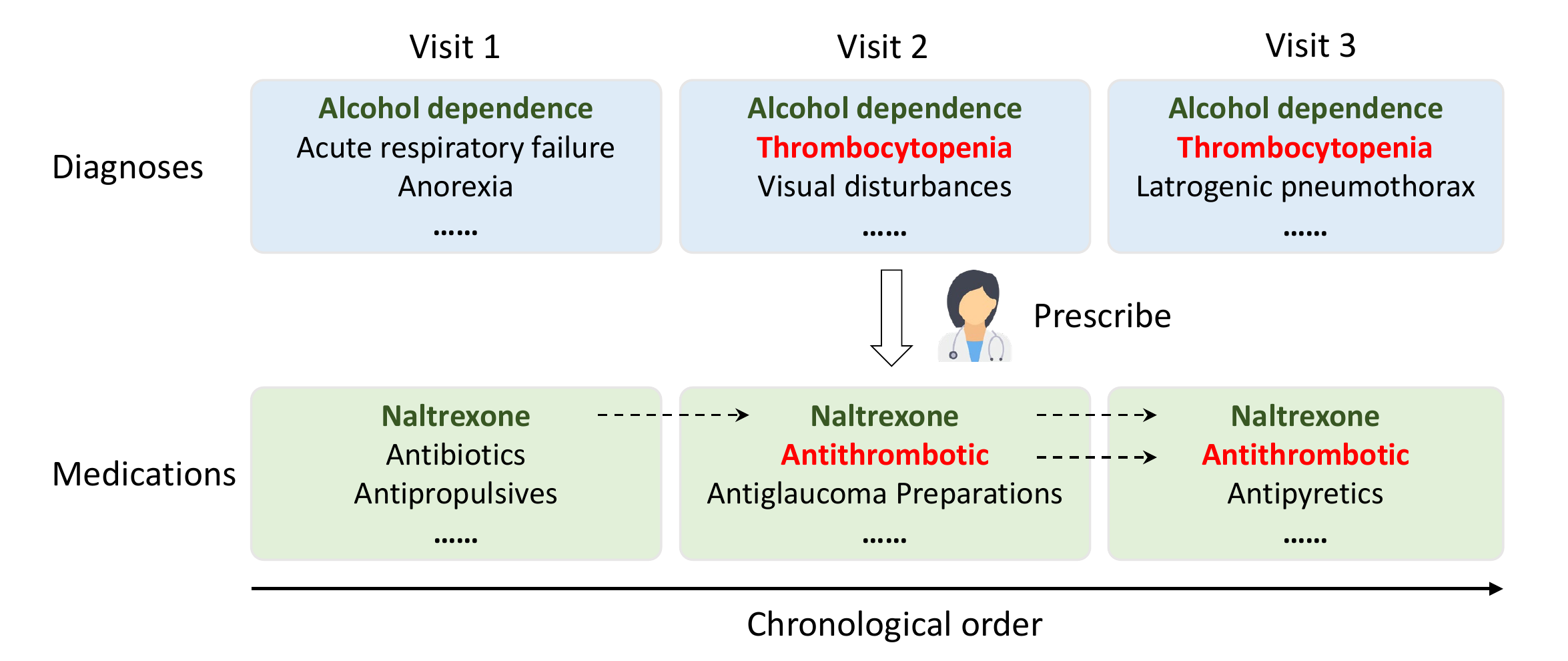}
	\caption{An example of electronic health record. \label{fig:data_sample}}
\end{figure}

\begin{figure}[h]
	\centering
	\includegraphics[width=\linewidth]{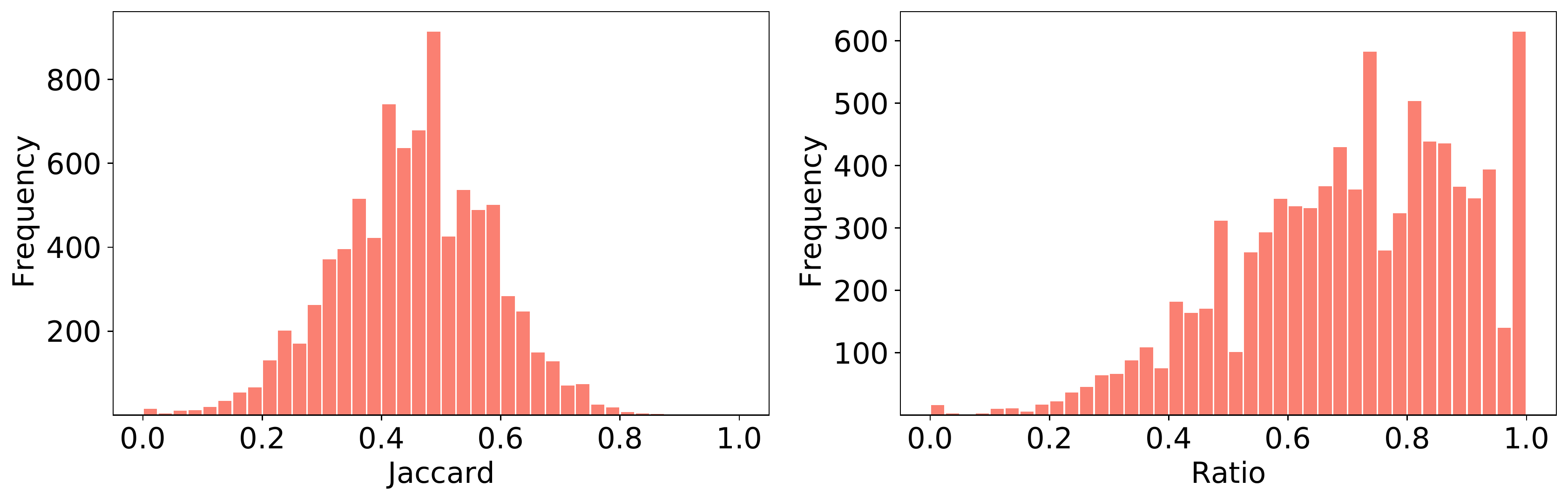}
	\caption{The histogram of Jaccard between current medications and historical medications (left) and the histogram of the proportion of current medications that occur in past visits (right). \label{fig:data_statistic}}
\end{figure}

Medication recommendation aims to provide a set of medicines to treat the set of diagnosed diseases of a patient. 
Take the patient in Figure~\ref{fig:data_sample} for example, this patient visits the hospital for three times. In each visit, this patient is diagnosed with a set of diseases and is prescribed a set of medicines to treat these diseases. Currently, medication recommendation is mainly conducted by doctors according to their expertise and experiences. However, many patients are diagnosed with multiple diseases at one time. To conduct a considerate medication recommendation, on one hand, the doctor needs to select proper medicines for each disease; on the other hand, the doctor needs to avoid harmful drug-drug interactions (DDI) among selected medicines. Therefore, for complicated cases, medication recommendation is time-consuming for experienced doctors and error-prone for inexperienced doctors. To address this problem, the automatic medication recommendation that can assist doctors in decision making is urged.


Due to the clinical value, medication recommendation has attracted growing research interests. A series of deep learning based medication recommendation methods \cite{safedrug, changematter} have been proposed, which can be divided into two categories: \textit{Instance-based} models \cite{LEAP, ecc, gong2021smr} only use patient's current diagnoses and procedures to conduct recommendations while ignoring the longitudinal patient history. In this manner, instance-based based models fail to consider the historical disease development process. To address this issue, \textit{Longitudinal} models \cite{safedrug, gamenet, changematter, dmnc} are designed to take use of the longitudinal patient history and capture the temporal dependencies. Existing longitudinal models usually consist of two stages, first aggregating the known information into a patient-level representation, and then conducting medication recommendation based on it. 

One problem of existing works is that they do not explicitly model the relationship between medication recommendations for the same patient. However, in clinical practice, the recommendations for the same patients are closely related. For example, for patients with chronic diseases, they may keep using the same medicine all their lives. As shown in Figure~\ref{fig:data_sample}, we conduct a statistical analysis on the MIMIC-III dataset. For each visit, we calculate the proportion of medications that have appeared in history and the Jaccard between current medications and past medications. We can see that in the most visits, a large portion of prescribed medicines have been recommended before. Inspired by this, we rethink about taking use of historical information from a medication-level perspective. The challenge here is how to accurately determine whether a historical medication is still relevant at present. 


In this paper, we propose an encoder-decoder based generation network to produce the appropriate medications in a sequential manner, named Conditional Generation Net (COGNet). The proposed COGNet consists of the basic model and the copy module. 
The basic model conducts recommendation only based on patients' health conditions in current visit; the copy module introduces the information of historical visits in modeling. Different from the basic model, in generating each medicine, the copy module decides whether to copy a medicine from historical recommendations or to predict a new one.
The experiments on a public dataset demonstrate the effectiveness of the proposed model. We summarize our major contributions as follows:
\begin{itemize}
    \item We propose a medication recommendation model, COGNet, which introduces a novel {\em copy-or-predict} mechanism. COGNet can leverage historical recommendations to produce a more accurate recommendation. 
    \item We develop a novel hierarchical selection mechanism, which chooses the reusable medicines to copy from both medication-level and visit-level perspectives. This increases the intepretability of the proposed COGNet.
    \item We conduct comprehensive experiments on a public dataset MIMIC-III to demonstrate the effectiveness of the proposed COGNet.
\end{itemize}


\section{Problem Formulation}

\subsection{Electrical Health Records (EHR)}
The basic unit of EHR is patient and each patient consists of several visits. Let $\mathcal{R} = {\{\mathcal{V}^{(i)}\}}_{i=1}^{N}$ denote a longitudinal EHR of $N$ patients. Each patient has visited the hospital as least once and the $i$-th patient can be represented as a sequence of multivariate observations  $\mathcal{V}^{(i)}=[\mathcal{V}_1^{(i)}, \mathcal{V}_2^{(i)}, \cdots, \mathcal{V}_{T^{(i)}}^{(i)}]$. Here $T^{(i)}$ is the number of visits of the $i$-th patient. 
To simplify the notation, we ignore the index $i$ and describe our methods with a single patient, then a patient is represented as $\mathcal{V}=[\mathcal{V}_1,\mathcal{V}_2,\cdots,\mathcal{V}_{T}]$. 
Let $\mathcal{D}=\{d_1, d_2, \cdots, d_{|\mathcal{D}|}\}$ denotes the set of diagnoses, $\mathcal{P}=\{p_1, p_2, \cdots, p_{|\mathcal{P}|}\}$ denotes the set of procedures and $\mathcal{M}=\{m_1, m_2, \cdots, m_{|\mathcal{M}|}\}$ denotes the set of medications. $|\mathcal{D}|$, $|\mathcal{P}|$ and $|\mathcal{M}|$ indicate the number of all possible diagnoses, procedures and medications, respectively.
Then, each visit of the patient can be represented by $\mathcal{V}_t = \{\mathcal{D}_t, \mathcal{P}_t, \mathcal{M}_t\}$, where $\mathcal{D}_t \subseteq \mathcal{D}$, $\mathcal{P}_t \subseteq \mathcal{P}$ and $\mathcal{M}_t \subseteq \mathcal{M}$.

\subsection{EHR\&DDI Graph}
$ G_e=\{\mathcal{M}, \mathcal{E}_e\} $ and $ G_d=\{\mathcal{M}, \mathcal{E}_d\} $ denote the EHR graph and DDI graph respectively, where $\mathcal{E}_e$ is all the possible medication combinations in $\mathcal{R}$ and $\mathcal{E}_d$ is the known DDIs. Formally, we use the adjacency matrix $A_e, A_d \in \mathbb{R}^{|\mathcal{M}|\times|\mathcal{M}|}$ to illustrate the construction of the graphs. $A_e[i,j]=1$ means the $i$-th and $j$-th medications have appeared in the same visit. For $A_d$, only the pair-wise DDIs are considered, $A_d[i,j]=1$ means the $i$-th and $j$-th medications are mutually interacted. $G_e$ and $G_d$ are the same for all patients.

\subsection{Medication Recommendation Problem}
Given a patient's current diagnoses $\mathcal{D}_t$, procedures $\mathcal{P}_t$, historical visit information $[\mathcal{V}_1, \mathcal{V}_2, \cdots, \mathcal{V}_{t-1}]$, and the EHR and DDI graphs $G_e$ and $G_d$, the goal is to train a model (i.e., COGNet) which can recommend the proper medication combination  $\mathcal{M}_t$ for this patient.

\begin{table}[h]
	\caption{Notations used in COGNet}
	\centering
	\label{notation}
		\begin{tabular}{l|l}
			\toprule
			\textbf{Notation}    & \textbf{Description} \\
			\midrule
			\midrule
			$\mathcal{R}$       & Electronic Health Records  \\
			$\mathcal{V}$      & Record for a single patient  \\
			$\mathcal{D}, \mathcal{P}, \mathcal{M}$      & Diagnoses, Procedure and  Medication set \\
			\midrule			
			$G_*$   & EHR or DDI Graph        \\
			$\mathcal{E_*}, A_*$     & Edge set and adjacency matrix of graph $G_*$        \\
			\midrule
			$\mathbf{E}_* \in \mathbb{R}^{|*| \times s} $ & Origin embedding tables \\
			$\mathbf{E}_g \in \mathbb{R}^{|\mathcal{M}| \times s} $ & Graph embedding for medications \\
			\bottomrule
		\end{tabular}
\end{table}

\section{Framework}


\begin{figure*}[h]
	\centering
	\includegraphics[width=0.85\linewidth]{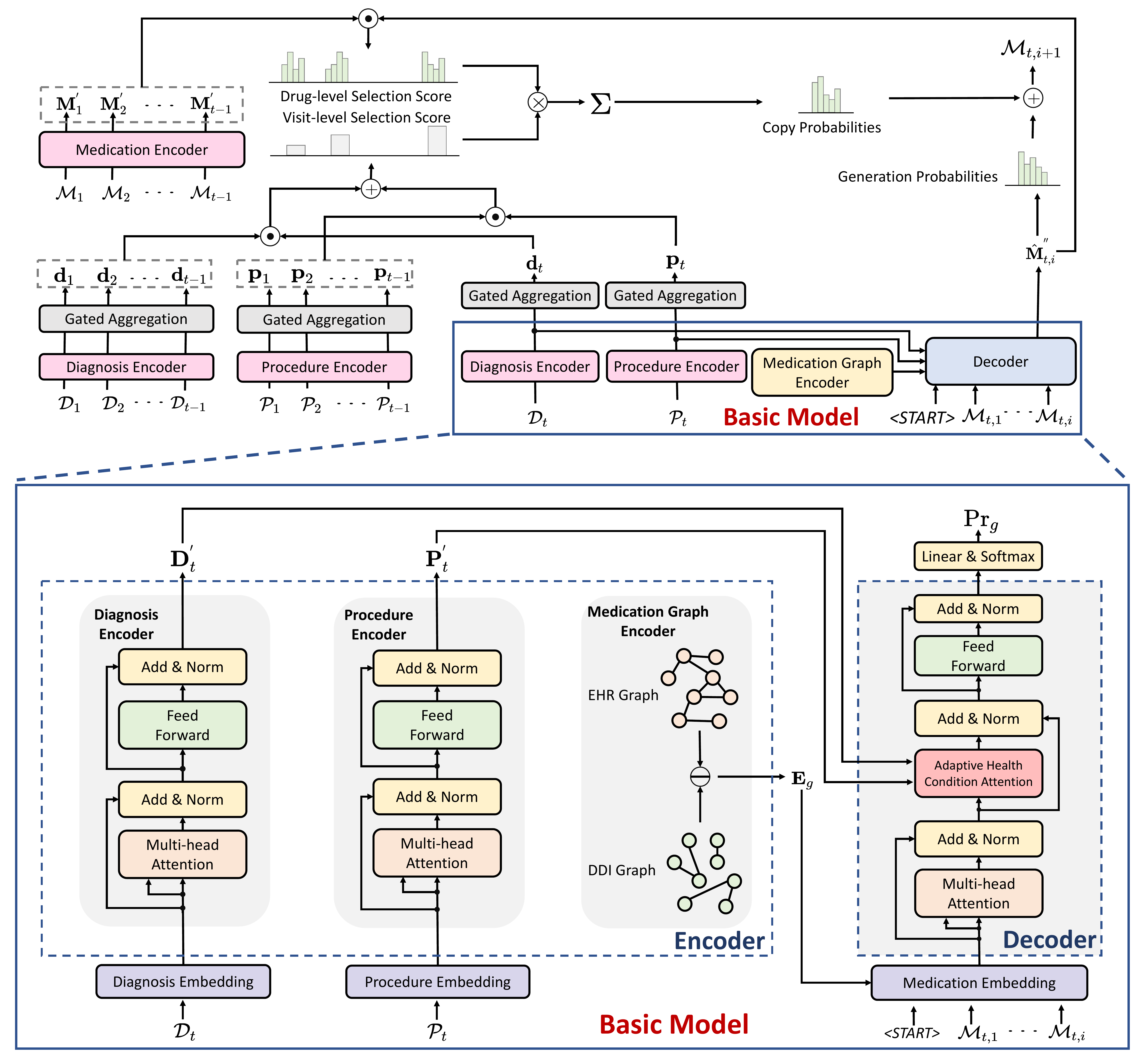}
	\caption{An overview of our proposed COGNet. The basic model recommend the medication only based on the patient's health condition in current visit. The other part, named copy module, considers the information of patient's historical visits. \label{fig:framework}}
\end{figure*}

Figure~\ref{fig:framework} illustrates the architecture of COGNet.
We employ an encoder-decoder based generative network to generate the appropriate medication combination based on the patient's current and historical health conditions.
In the encoder, we encode all medical codes of the historical visits (i.e., $\mathcal{V}_1$,$\mathcal{V}_2$,$\cdots$,$\mathcal{V}_{t-1}$) to represent the patient's historical health conditions and encode the diagnosis and procedure codes of the $t-$th visit to represent the patient's current health condition.
In the decoder, we will generate the medication one by one to compose the recommendation drug combination.
During each decoding step, the decoder combines the information of the diagnoses, procedures and generated medications to recommend the next medication.
Moreover, if some diseases in current visit keep consistent with the past visits, the copy module will directly copy the corresponding medications from historical drug combinations.
We will first describe the basic encoder-decoder based generation model and then address the copy module in detail in the following subsections.

\subsection{Input Representations}
We design three embedding tables, $\mathbf{E}_d \in \mathbb{R}^{|\mathcal{D}| \times s}$, $\mathbf{E}_p \in  \mathbb{R}^{|\mathcal{P}| \times s}$ and $\mathbf{E}_m \in  \mathbb{R}^{|\mathcal{M}| \times s}$, where each row is an embedding vector for a distinct diagnosis code, procedure code or medication code. Here $s$ denotes the dimension of the embedding space. 
For each diagnosis set $\mathcal{D}_i$ ($i\in [1, t]$), we first convert its every element $d\in \mathcal{D}_i$ to a $s$-dimensional vector $\mathbf{d}$ via the embedding matrix $\mathbf{E}_d$.
Then, we can obtain the representation of the diagnosis set $\mathbf{D}_i \in \mathbb{R}^{|\mathcal{D}_t| \times s}$.
For each procedure set $\mathcal{P}_i$ ($t\in [1, t]$) and medication set $\mathcal{M}_j$ ($j\in [1, t-1]$), we can also obtain their representations $\mathbf{P}_i \in \mathbb{R}^{|\mathcal{P}_i| \times s}$ and $\mathbf{M}_j \in \mathbb{R}^{|\mathcal{M}_j| \times s}$ via the embedding matrices $\mathbf{E}_p$ and $\mathbf{E}_m$, respectively.


\subsection{Basic Model}
In the basic model, we recommend the medication combination only based on the patient's health condition in current visit.
The basic model is an encoder-decoder generation model.
It consists of four modules: \textit{diagnosis encoder}, \textit{procedure encoder}, \textit{medication graph encoder}, and \textit{medication combination decoder}.

\subsubsection{Diagnosis Encoder}
The diagnosis encoder aims to represent the patient's health conditions based on the diagnosis codes.
It is a Transformer-based~\cite{transformer} network and has the following two major sub-layers.

\textit{Multi-Head Self-Attention}. This sub-layer aims to capture relations among all diagnoses in the same visit. 
Given three input matrices \(\mathbf{Q}\in \mathbb{R}^{L_Q\times s}\), \(\mathbf{K}\in \mathbb{R}^{L_K\times s}\) and \(\mathbf{V}\in  \mathbb{R}^{L_V\times s}\) where \(L_K=L_V\), the attention function is defined as:
\begin{equation}
\operatorname{Attention}(\mathbf{Q}, \mathbf{K}, \mathbf{V}) 
           = \operatorname{Softmax}(\frac{\mathbf{Q}\mathbf{K}^\top}{\sqrt{s}}) \mathbf{V}
\end{equation}
Multi-head attention layer \(\text{MH}(\cdot,\cdot,\cdot)\) will further project the input to multiple representation subspaces and capture the interaction information from multiple views~\cite{DBLP:conf/aaai/QiuWG021}.  
\begin{equation}
\begin{split}
    & \operatorname{MH}(\mathbf{Q}, \mathbf{K}, \mathbf{V}) = [\operatorname{head}_1;...;\operatorname{head}_h]\mathbf{W}^O \\
    & \operatorname{head}_i = \operatorname{Attention}(\mathbf{Q} \mathbf{W}^Q_i, \mathbf{K} \mathbf{W}^K_i, \mathbf{V} \mathbf{W}^V_i)
\end{split}
\end{equation}
\(\mathbf{W}^Q_i\), \(\mathbf{W}^K_i\), \(\mathbf{W}^V_i\in \mathbb{R}^{s\times s/h}\) and \(\mathbf{W}^O\in \mathbb{R}^{s\times s}\) are the parameters to learn. \(h\) is the number of heads.

\textit{Position-wise Feed-Forward}. 
This sub-layer consists of two linear projections with a \(\textit{ReLU}\) activation in between.
\begin{equation}
    \operatorname{FFN}(\mathbf{H}) = \operatorname{ReLU}(\mathbf{H} \mathbf{W}^F_1 + \mathbf{b}^F_1)\mathbf{W}^F_2 + \mathbf{b}^F_2
\end{equation}
where \(\mathbf{W}^F_1\in \mathbb{R}^{s\times 8s}\), \(\mathbf{W}^F_2\in \mathbb{R}^{8s\times s}\), \(\mathbf{b}^F_1\in \mathbb{R}^{8s}\) and \(\mathbf{b}^F_2\in \mathbb{R}^{s}\)
are trainable parameters. 

The diagnosis encoder then employs the residual connection and layer normalization function \(\operatorname{LayerNorm}(\cdot)\) defined in ~\cite{ba2016layer} around the above two sub-layers to extract the diagnosis representation:
\begin{equation}
    \begin{split}
        \mathbf{D}^{\prime}_t &= \operatorname{Enc_d}(\mathbf{D}_t) = \operatorname{LayerNorm}(\mathbf{H} + \operatorname{FFN}(\mathbf{H})) \\
      \text{where}\ \   \mathbf{H} &= \operatorname{LayerNorm}(\mathbf{D}_t + \operatorname{MH}(\mathbf{D}_t,\mathbf{D}_t,\mathbf{D}_t))
    \end{split}
\end{equation}
where $\mathbf{D}^{\prime}_t \in \mathbb{R}^{|\mathcal{D}_t| \times s}$ is the diagnosis representation of the $t$-th visit encoded by this encoder.

\subsubsection{Procedure Encoder}
The procedure encoder has the same network structure as the diagnosis encoder but their parameters are different.
Through the same encoding process, we can also obtain the procedure representation $\mathbf{P}^{\prime}_t$ via the procedure encoder:
\begin{equation}
    \mathbf{P}^{\prime}_t = \operatorname{Enc_p}(\mathbf{P}_t)
\end{equation}

\subsubsection{Medication Graph Encoder}
This encoder aims to model two kinds of drug relations:
\begin{itemize}
    \item \textit{Medication EHR co-occurrence relationship}: Some medications are often prescribed together for better efficacy. For example, ranitidine and sucralfate are the popular drug combination used to treat acute gastritis. Hence, modeling these co-occurrence relations can help the decoder to recommend the drug according to the partially generated medications.
    \item \textit{DDI}: As mentioned above, some medications have the DDI and can not be used together. When recommending the drug, the decoder should avoid that it is conflict with the past recommended drugs. Hence, modeling this relation can help to recommend a safe and effective medication combination.
\end{itemize}
Inspired by ~\cite{gamenet}, we use the Graph Convolutional Network~\cite{gcn} to model two kinds of relations based on the medication EHR and DDI graphs.

Given the input medication features $\mathbf{X}\in \mathbb{R}^{|\mathcal{M}|\times s}$ and the medication graph adjacency matrix $\mathbf{A}\in \mathbb{R}^{|\mathcal{M}|\times |\mathcal{M}|}$, the $\operatorname{GCN}(\cdot, \cdot)$ will obtain the new medication representations as follows:
\begin{equation}
    \operatorname{GCN}(\mathbf{X}, \mathbf{A}) = \sigma(\hat{\mathbf{O}}^{-\frac{1}{2}} \hat{\mathbf{A}} \hat{\mathbf{O}}^{-\frac{1}{2}} \mathbf{X})
\end{equation}
$\hat{\mathbf{A}} = \mathbf{A}+\mathbf{I}$, $\mathbf{I}$ is the identity matrix and $\hat{\mathbf{O}}$ is the diagonal node degree matrix of $\hat{\mathbf{A}}$ (i.e., $\mathbf{O}_{i,i} = \sum_j \mathbf{A}_{i,j}$).

First, we use a two-layer GCN to model the medication EHR co-occurrence relations based on the EHR graph adjacency matrix $\mathbf{A}_e$.
\begin{equation}
    \mathbf{G}_e = \operatorname{GCN}(\operatorname{ReLU}(\operatorname{GCN}(\mathbf{E}_m, \mathbf{A}_e))\mathbf{W}^g_{e}, \mathbf{A}_e)
\end{equation}
where $\mathbf{W}^g_{e}$ is the learnable parameter.

Then, we use another two-layer GCN to model the DDI relations based on the DDI graph $\mathbf{A}_d$.
\begin{equation}
    \mathbf{G}_d = \operatorname{GCN}(\operatorname{ReLU}(\operatorname{GCN}(\mathbf{E}_m, \mathbf{A}_d))\mathbf{W}^g_{d}, \mathbf{A}_d)
\end{equation}
where $\mathbf{W}^g_{d}$ is the learnable parameter.

Finally, we fuse two medication relation representations $\mathbf{G}_e$ and $\mathbf{G}_d$ to obtain the relation-aware medication representations.
\begin{equation}
    \mathbf{E}_g = \mathbf{G}_e - \lambda \mathbf{G}_d
\end{equation}
where $\lambda$ is a learnable parameter.

\subsubsection{Medication Combination Decoder}
The decoder will recommend the medication one by one for the $t-$th visit.
For example, at the $i$-th decoding step, the decoder will recommend the next medication $\mathcal{M}_{t,i}$ based on the partially generated medication combination $\{\mathcal{M}_{t,0},\cdots, \mathcal{M}_{t,i-1}\}$ and the patient health conditions in the $t$-th visit (i.e., $\mathbf{D}^{\prime}_t$ and $\mathbf{P}^{\prime}_t$).
The decoding process is as follows.

First, we convert all generated medications to vectors via the original embedding matrix $\mathbf{E}_m$ and relation representation matrix $\mathbf{E}_g$ and can obtain two representations of the partially generated medication combination $\hat{\mathbf{M}}_t^m$ and $\hat{\mathbf{M}}_t^g$.
We fuse two representations to obtain the relation-aware medication combination representation.
\begin{equation}
\label{eq:med emb combine}
\hat{\mathbf{M}}_t = \hat{\mathbf{M}}_t^m + \hat{\mathbf{M}}_t^g
\end{equation}

Then, we use the multi-head self-attention mechanism (defined in Eq.(2)) to capture the interactions among the recommended medications.
\begin{equation}
    \hat{\mathbf{M}}_t^{'} = \operatorname{LayerNorm}( \hat{\mathbf{M}}_t + \operatorname{MH}(\hat{\mathbf{M}}_t, \hat{\mathbf{M}}_t, \hat{\mathbf{M}}_t))
\end{equation}

Intuitively, the medication recommendation task aims to generate the recommendation of drugs that can cover all diseases of the patient.
Thus, the patient's disease and procedure representations $\mathbf{D}_t^{'}$ and $\mathbf{P}_t^{'}$ play an auxiliary role during the decoding process.
To this end, we align the patient's health conditions to adaptively model the uncovered diseases to guide the next medication recommendation.
\begin{equation}
    \hat{\mathbf{M}}_t^{''} = \operatorname{LayerNorm}( \hat{\mathbf{M}}_t^{'} + \operatorname{MH}(\hat{\mathbf{M}}_t^{'}, \mathbf{D}_t^{'}, \mathbf{D}_t^{'}) + \operatorname{MH}(\hat{\mathbf{M}}_t^{'}, \mathbf{P}_t^{'}, \mathbf{P}_t^{'}))
\end{equation}

Finally, we use the last row of $\hat{\mathbf{M}}_t^{''}$ (i.e., $\hat{\mathbf{M}}_{t,i-1}^{''}$) to predict the $i$-th medication via an MLP layer.
\begin{equation}
    {\Pr}_g = \operatorname{Softmax}(\hat{\mathbf{M}}_{t,i-1}^{''} \textbf{W}_g + \textbf{b}_g)
\end{equation}
where $\textbf{W}_g\in \mathbb{R}^{s\times |\mathcal{M}|}$ and $\textbf{b}_g\in \mathbb{R}^{|\mathcal{M}|}$ are learnable parameters.
${\Pr}_g$ denotes the probabilities of all medications in the vocabulary in which the medication with the maximum probability is the predicted medication at step $i$.

\subsection{Copy Module}
In \textit{Basic Model} section, we introduce the encoder-decoder based model to generate the medication recommendation results based on the patient's current health conditions.
It doesn't consider the information of patient's historical visits.
In this section, we design a novel copy module to extend the basic model, which first compares the health conditions of current and historical visits and then copies the reusable medications to prescribe for current visit according to the condition changes.

Since the patient may have multiple visits, we use the hierarchical selection mechanism to conduct the copy process at each decoding step.
First, we use the visit-level selection to pick a similar visit by comparing their health conditions.
Then, at the specific decoding step, we use the medication-level selection to pick a particular medication from the prescriptions of the selected visit and add it to the recommendation results.
Finally, we repeat the above process to copy  the reusable medications iteratively to form the complete recommendation medication list.
Note that the selection process is ``soft'' like the attention mechanism~\cite{DBLP:conf/icml/XuBKCCSZB15} and assigns different probabilities for all visits/medications to highlight the choice.
We will take the $i$-th medication recommendation of the $t$-th visit (i.e., $\mathcal{M}_{t,i}$) as an example to illustrate the copy process.

\subsubsection{Visit-level Selection}
We first uses two gated aggregation layers to encode the visit-level health conditions of all visits by aggregating their diagnosis and procedure representations, respectively:
\begin{equation}
    \mathbf{v}^d_j = \operatorname{Softmax}({\tanh(\mathbf{D}^{\prime}_j \mathbf{W}^{1}_{d} + \mathbf{b}^1_d)\mathbf{W}^{2}_{d} + b^2_d})^\top {\mathbf{D}^{\prime}_j}
\end{equation}
\begin{equation}
    \mathbf{v}^p_j = \operatorname{Softmax}({\tanh(\mathbf{P}^{\prime}_j \mathbf{W}^{1}_{p} + \mathbf{b}^1_p)\mathbf{W}^{2}_{p} + b^2_p})^\top {\mathbf{P}^{\prime}_j}
\end{equation}
where $\mathbf{W}^1_* \in \mathbb{R}^{s \times \delta}$, $\mathbf{b}^1_* \in \mathbb{R}^{\delta}$, $\mathbf{W}^2_* \in \mathbb{R}^{\delta \times 1}$ and $b^2_* \in \mathbb{R}$ are trainable parameters. Then we calculate the visit-level selection score of the past $j$-th visit ($1\le j \le t-1$) by measuring the similarity between it and the current $t$-th visit.
\begin{equation}
    c_j = \operatorname{Softmax}({\frac{\mathbf{v}^d_j \cdot \mathbf{v}^d_t + \mathbf{v}^p_j \cdot \mathbf{v}^p_t}{\sqrt{s}}})
\end{equation}
$c_j \in \mathbb{R}$ denotes the selection score of the $j$-th visit.

\subsubsection{Medication-level Selection}
We use the hidden state $\mathbf{M}^{''}_{t,i-1}$ formulated in Eq.(12), which comprehensively encodes the information of diagnoses, procedures and recommended medications, to determine which historical medications $m\in \bigcup_{j=1}^{t-1} \mathcal{M}_{j}$ are reusable in current situation.

We firstly obtain the medication representations of all past visits through the similar process in Section 3.2.1:
\begin{equation}
	\mathbf{M}^{'}_j = \operatorname{Enc_m}(\mathbf{M}_j)
\end{equation}
where $\mathbf{M}^{'}_j \in \mathbb{R}^{|\mathcal{M}_j| \times s}$. 

Then, we use the hidden state $\hat{\mathbf{M}}_{t,i-1}^{''}$ as the query vector to calculate the selection score along the medication dimension.
The selection score of $k$-th medication in $j$-th visit $\mathcal{M}_{j,k}$ ($1\le j \le t-1$ and $1\le k \le |\mathcal{M}_j|$) is
\begin{gather}
q_{j,k} = \frac{\exp (\hat{q}_{j,k})}{\sum_{j=1}^{t-1} \sum_{k=1}^{|\mathcal{M}_j|} \exp (\hat{q}_{j,k})} \\
\hat{q}_{j,k} = \frac{ (\hat{\mathbf{M}}_{t,i-1}^{''} \textbf{W}_c) \cdot \mathbf{M}^{'}_{j,k}}{\sqrt{s}}
\end{gather}
where $\textbf{W}_c \in \mathbb{R}^{s \times s}$ is a learnable parameter. 
$q_{j,k} \in \mathbb{R}$ denotes the medication-level selection score.

\subsubsection{Copy Mechanism}
We combine the visit-level and medication-level scores to determine the copy probability of each medication in past visits.
Moreover, since a medication $m_i\in \mathcal{M}$ may be used in multiple past prescriptions, we gather its final copy probability as follows: 
\begin{gather}
    p^{(i)}_c = \frac{\hat{p}^{(i)}_c}{\sum_{i=1}^{|\mathcal{M}|} \hat{p}^{(i)}_c} \\
    \textit{where} \ \  \hat{p}^{(i)}_c = \sum_{j=1}^{t-1} \sum_{k=1}^{|\mathcal{M}_j|} q_{j,k} * c_j * \mathbf{1}_{\{\mathcal{M}_{j,k}=m_i\}}
\end{gather}
where $\mathbf{1}_{\{\mathcal{M}_{j,k}=m_i\}}$ is an indicator function which returns 1 if $\mathcal{M}_{j,k}=m_i$,
and 0 if $\mathcal{M}_{j,k}\ne m_i$.
The copy probabilities of all medications is ${\Pr}_c=[p^{(1)}_c,p^{(2)}_c, ... ,p^{(|\mathcal{M}|)}_c]\in \mathbb{R}^{|\mathcal{M}|}$.

Finally, we combine the generation probabilities and copy probabilities to conduct the prediction.
\begin{gather}
    \label{eq:prediction}
    \Pr = w_g * {\Pr}_g + (1-w_g)*{\Pr}_c \\
    w_g = \operatorname{Sigmoid}(\hat{\mathbf{M}}^{''}_{t,i-1} \mathbf{W}_f + b_f)
\end{gather}
where $\mathbf{W}_f \in \mathbb{R}^{s\times 1}$ and $b_f\in \mathbb{R}$ are learnable parameters.
$\Pr$ denotes the probabilities of all medications in the medication set $\mathcal{M}$ in which the medication with the maximum probability is the predicted $i$-th medication of the $t-$th visit.
	
\subsection{Training}
We train the model by minimizing regular cross-entropy loss:
\begin{equation}
    \mathcal{L}(\theta) = -\sum_{\mathcal{R}}\sum_{t}\sum_{i}\log \Pr (\mathcal{M}_{t,i}|\mathcal{V}_{<t},\mathcal{D}_t,\mathcal{P}_t,\mathcal{M}_{t,<i};\theta)
\end{equation}
where $\mathcal{R}$ is the training EHR set.
$\mathcal{M}_{t,i}$ is the $i$-th medication of the medication set of the $t-$th visit.
$\Pr (\mathcal{M}_{t,i}|\mathcal{V}_{<t},\mathcal{D}_t,\mathcal{P}_t,\mathcal{M}_{t,<i};\theta)$ is the predicted probability of the $\mathcal{M}_{t,i}$ and can be calculated by the Eq.(\ref{eq:prediction}). 
$\theta$ denotes all trainable parameters in COGNet.

During the training phase, we use the teacher-forcing to train the model.
Specifically, when predicting the $i$-th medication, the model takes the real $(i-1)$-th medication rather than the predicted $(i-1)$-th medication as the input of the decoder.
At the first step, the input of the decoder is set to a special token $\langle \textit{START}\rangle$.

\begin{table*}[ht]
	\caption{Performance Comparison on MIMIC-III. Best results are highlighted in bold. }
	\centering
	\label{performance}
		\begin{tabular}{l|ccccc}
			\toprule
			Model    & Jaccard & F1 & PRAUC & DDI   & Avg. \# of Drugs \\
			\midrule
			\midrule
			LR       & 0.4865 $\pm$ 0.0021 & 0.6434 $\pm$ 0.0019 & 0.7509 $\pm$ 0.0018 & 0.0829 $\pm$ 0.0009 & 16.1773 $\pm$ 0.0942\\
			ECC      & 0.4996 $\pm$ 0.0049 & 0.6569 $\pm$ 0.0044 & 0.6844 $\pm$ 0.0038 & 0.0846 $\pm$ 0.0018 & 18.0722 $\pm$ 0.1914\\
			\midrule			
			RETAIN   & 0.4887 $\pm$ 0.0028 & 0.6481 $\pm$ 0.0027 & 0.7556 $\pm$ 0.0033 & 0.0835 $\pm$ 0.0020 & 20.4051 $\pm$ 0.2832\\
			LEAP     & 0.4521 $\pm$ 0.0024 & 0.6138 $\pm$ 0.0026 & 0.6549 $\pm$ 0.0033 & 0.0731 $\pm$ 0.0008 & 18.7138 $\pm$ 0.0666\\
			DMNC     & 0.4864 $\pm$ 0.0025 & 0.6529 $\pm$ 0.0030 & 0.7580 $\pm$ 0.0039 & 0.0842 $\pm$ 0.0011 & 20.0000 $\pm$ 0.0000\\
			GAMENet  & 0.5067 $\pm$ 0.0025 & 0.6626 $\pm$ 0.0025 & 0.7631 $\pm$ 0.0030 & 0.0864 $\pm$ 0.0006 & 27.2145 $\pm$ 0.1141\\
			MICRON   & 0.5100 $\pm$ 0.0033 & 0.6654 $\pm$ 0.0031 & 0.7687 $\pm$ 0.0026 & 0.0641 $\pm$ 0.0007 & 17.9267 $\pm$ 0.2172\\
			SafeDrug & 0.5213 $\pm$ 0.0030 & 0.6768 $\pm$ 0.0027 & 0.7647 $\pm$ 0.0025 & \textbf{0.0589 $\pm$ 0.0005} & 19.9178 $\pm$ 0.1604\\
			\midrule
			\textbf{COGNet} & \textbf{0.5336 $\pm$ 0.0011} & \textbf{0.6869 $\pm$ 0.0010} & \textbf{0.7739 $\pm$ 0.0009} & 0.0852 $\pm$ 0.0005 & 28.0903 $\pm$ 0.0950 \\
			\bottomrule
		\end{tabular}
\end{table*}

\begin{table}[ht]
	\caption{Statistics of processed MIMIC-III. \label{fig:data_stats}}
	\centering
	\label{statistic}
		\begin{tabular}{l|l}
			\toprule
			\textbf{Item}    & \textbf{Number} \\
			\midrule
			\midrule
			\# of visits / \# of patients & 14,995 / 6,350 \\
            diag. / prod. / med. space size & 1,958 / 1430 / 131\\
            avg. / max \# of visits & 2.37 / 29\\
            avg. / max \# of diagnoses per visit & 10.51 / 128\\
            avg. / max \# of procedures per visit & 3.84 / 50\\
            avg. / max \# of medications per visit & 11.44 / 65\\
            total \# of DDI pairs & 448\\
			\bottomrule
		\end{tabular}
\end{table}

\subsection{Inference}
During the inference phase, inspired by the medication recommendation method~\cite{LEAP} and some NLP generation methods~\cite{DBLP:conf/nlpcc/ZhouYWTBZ17,DBLP:conf/emnlp/NemaMKSR19,DBLP:conf/coling/QiuWF20}, we use the beam search trick to improve the performance.
Different from the greedy search which only selects one best candidate as the partially generated medication combination for each decoding step, the beam search algorithm selects multiple alternatives at each decoding step based on conditional probability.

\section{Experiments}
In this section, we first introduce the experimental setups. Then we conduct some experiments to demonstrate the effectiveness of our COGNet model\footnote{https://github.com/BarryRun/COGNet}.

\subsection{Dataset}
We use the Medical Information Mart for Intensive Care (MIMIC-III)\footnote{https://mimic.physionet.org/}\cite{Johnson2016MIMICIIIAF} dataset released on PhysioNet. 
It contains a total of 46520 patients and 58976 hospital admissions from 2001 to 2012. 
For a fair comparison, we use the data processing script \footnote{https://github.com/ycq091044/SafeDrug} released by \cite{safedrug} and take the processed data as our benchmark. Details of the processing can be found in Appendix.
Table~\ref{fig:data_stats} shows some statistics on the processed data.

\begin{figure*}[t]
    \centering
    \includegraphics[width=\linewidth]{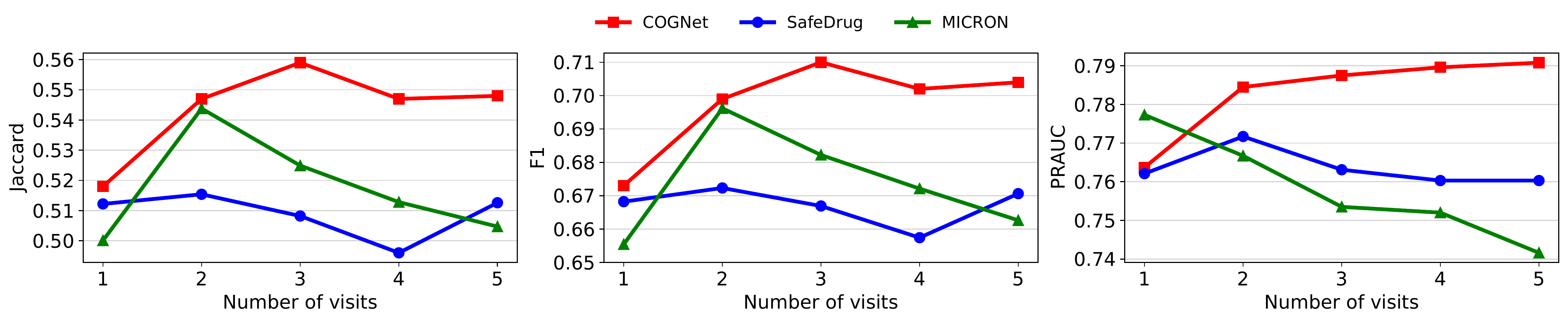}
    \caption{The line charts show the effect of number of visits for various models. }
    \label{fig:visits}
\end{figure*}

\subsection{Baselines and Metrics}

We evaluate the performance of our approach by comparing it with several baseline methods.
\begin{itemize}
    \item \textbf{LR}, standard Logistic Regression;
    \item \textbf{ECC}~\cite{ecc}, Ensemabled Classifier Chain, which uses a multi-hot vector to encode the diagnosis and procedure sets and leverage multiple SVM classifiers to make multi-label prediction;
    \item \textbf{RETAIN}~\cite{retain}, which uses the attention and gate mechanism to improve the prediction interpretability;
    \item \textbf{LEAP}~\cite{LEAP}, which uses the LSTM based generation model to conduct medication recommendation based on the diagnose information; 
    \item \textbf{DMNC}~\cite{dmnc}, which proposes a new memory augmented neural network model to improve the patient encoder;
    \item \textbf{GAMENet}~\cite{gamenet}, which further combines the memory network and graph neural network to recommend the medication combinations;
    \item \textbf{MICRON}~\cite{changematter}, which first uses an recurrent residual learning model to predict the medication changes and then conducts medication recommendation based on the medication changes and the medication combination of the last visit;
    \item \textbf{SafeDrug}~\cite{safedrug}, which combines the drug molecular graph and DDI graph to predict the safe medication combination.
\end{itemize}
Following the previous medication recommendation work~\cite{gamenet,safedrug,changematter}, we use Jaccard Similarity Score (Jaccard), Average F1 (F1), Precision Recall AUC (PRAUC), and DDI Rate as our evaluation metrics. Each metric is averaged over all patients. The metric definitions can be found in Appendix.

\begin{table*}
	\caption{Ablation Study for Different Components of COGNet on MIMIC-III.}
	\centering
	\label{ablation}
		\begin{tabular}{l|ccccc}
			\toprule
			Model    & Jaccard & F1 & PRAUC & DDI    & Avg. \# of Drugs \\
			\midrule			
			\midrule			
			COGNet \textit{w/o Copy}   & 0.5163 $\pm$ 0.0010 & 0.6713 $\pm$ 0.0009 & 0.7637 $\pm$ 0.0018 & 0.0842 $\pm$ 0.0005  &28.3139 $\pm$ 0.0766 \\
			COGNet \textit{w/o $c_i$}   & 0.5119 $\pm$ 0.0016 & 0.6629 $\pm$ 0.0014 & 0.7588 $\pm$ 0.0014 & 0.0813 $\pm$ 0.0005 & 26.8944 $\pm$ 0.0953 \\
			COGNet \textit{w/o $G$}     & 0.5306 $\pm$ 0.0013 & 0.6836 $\pm$ 0.0012 & 0.7706 $\pm$ 0.0013 & 0.0840 $\pm$ 0.0002 & 29.1076 $\pm$ 0.0795 \\
			COGNet \textit{w/o $\mathcal{D}$}  & 0.4937 $\pm$ 0.0011 & 0.6496 $\pm$ 0.0011 & 0.7443 $\pm$ 0.0014 & 0.0887 $\pm$ 0.0004 & 28.0519 $\pm$ 0.0995\\
			COGNet \textit{w/o $\mathcal{P}$}  & 0.5117 $\pm$ 0.0010 & 0.6669 $\pm$ 0.0010 & 0.7625 $\pm$ 0.0016 & 0.0831 $\pm$ 0.0002 & 28.9554 $\pm$ 0.0885 \\
			COGNet \textit{w/o BS} & 0.5266 $\pm$ 0.0021 & 0.6805 $\pm$ 0.0019 & 0.7729 $\pm$ 0.0013 & 0.0840 $\pm$ 0.0004 & 28.5592 $\pm$ 0.0701 \\
			\midrule
			\textbf{COGNet} & \textbf{0.5336 $\pm$ 0.0011} & \textbf{0.6869 $\pm$ 0.0010} & \textbf{0.7739 $\pm$ 0.0009} & 0.0852 $\pm$ 0.0005 & 28.0903 $\pm$ 0.0950 \\
			\bottomrule
		\end{tabular}
\end{table*}

\subsection{Result Analysis}

\subsubsection{Overall Comparison}
Table~\ref{performance} shows the results of all methods. Overall, our proposed model COGNet outperforms all baselines with the higher Jaccard, F1 and PRAUC.
The performances of LR, ECC and LEAP are poor as they are instance-based models that only consider the diagnoses and procedures in the current visit.  
RETAIN, DMNC, GAMENet, SafeDrug and MICRON perform relatively better because they preserve longitudinal patient information in different ways. 
RETAIN and DMNC only encode the patients' historical information, while GAMENet introduces additional graph information and SafeDrug incorporates the drug molecule structures in medication recommendation, resulting in a further performance improvement. 
MICRON also notices that some medications in the current visit keeps consistent with the last visit and uses the recurrent residual method to inherit the unchanged part.
However, it fails to consider the correlations among the recommended medications and the recurrent like method is hard to model the long range visit information.
Hence, COGNet performs better than MICRON.
SafeDrug achieves a lower DDI rate by introducing the additional drug molecule information.
However, the MIMIC-III dataset itself has an average DDI of 0.08379 and our COGNet has a similar performance.
It suggests that COGNet mimics the behavior of physicians in prescribing medications well.

\subsubsection{Effect of number of visits}
To further explore whether our COGNet can better capture historical medication information, we investigate the impact of the number of visits on the performance of different models. 
Since most patients in MIMIC visit the hospital less than five times, we take the first five visits for each patient in the test set to conduct the analysis. 
As a comparison, we also take the two strongest baselines SafeDrug and MICRON, which also incorporate historical information, to conduct the analysis. 
Figure~\ref{fig:visits} shows the results. 
We can see that COGNet achieves relatively better performance with more visits, while the performance of SafeDrug almost stays flat and MICRON shows a decreasing trend. 
The reason may be that COGNet uses the attention based hierarchical selection mechanism, which can more effectively incorporate the information of past visits than RNN like mechanism used in SafeDrug.
Moreover, MICRON iteratively updates the past medication combination to form the new medication set, which will be affected by the error accumulation problem.


\subsection{Ablation Study}
To verify the effectiveness of each module of COGNet, we design the following ablation models:
\begin{itemize}
    \item COGNet \textit{w/o Copy}: We remove the copy module, which means changing the Eq.(\ref{eq:prediction}) to $\Pr = \Pr_g$.
    \item COGNet \textit{w/o $c_i$}: We maintain the copy module but remove the visit-level selection by changing the $q_{j,k} c_j$ in Eq.(21) to $q_{j,k}$.
    \item COGNet \textit{w/o $G$}: We remove the EHR and DDI graphs in decoder, which means changing the Eq.(\ref{eq:med emb combine}) to $\hat{\mathbf{M}}_t = \hat{\mathbf{M}}_t^m$.
    \item COGNet \textit{w/o $\mathcal{D}$}: We remove the diagnoses information in each visit.
    \item COGNet \textit{w/o $\mathcal{P}$}: We remove the procedures information in each visit.
    \item COGNet \textit{w/o BS}. We use the greedy search strategy in the inference phase rather than beam search.
\end{itemize}

Table~\ref{ablation} shows the results for the different variants of COGNet. 
As expected, the results of COGNet \textit{w/o Copy} indicate that the copy mechanism brings a significant improvement to the basic model. 
COGNet can improve the medication recommendation by replicating historical drugs. 
COGNet \textit{w/o $c_i$} illustrates the effectiveness of visit-level selection scores. 

Both COGNet \textit{w/o $\mathcal{D}$} and COGNet \textit{w/o $\mathcal{P}$} yield poor results among all ablation models, which suggest that diagnosis and procedure information play a great role in medication recommendation. The results of COGNet \textit{w/o $G$} and COGNet \textit{w/o BS} indicate that graphs and beam search also have contributions to the final result. Overall, the complete COGNet outperforms all ablation models, which means each component of our model is integral.

\begin{figure}
    \centering
    \includegraphics[width=\linewidth]{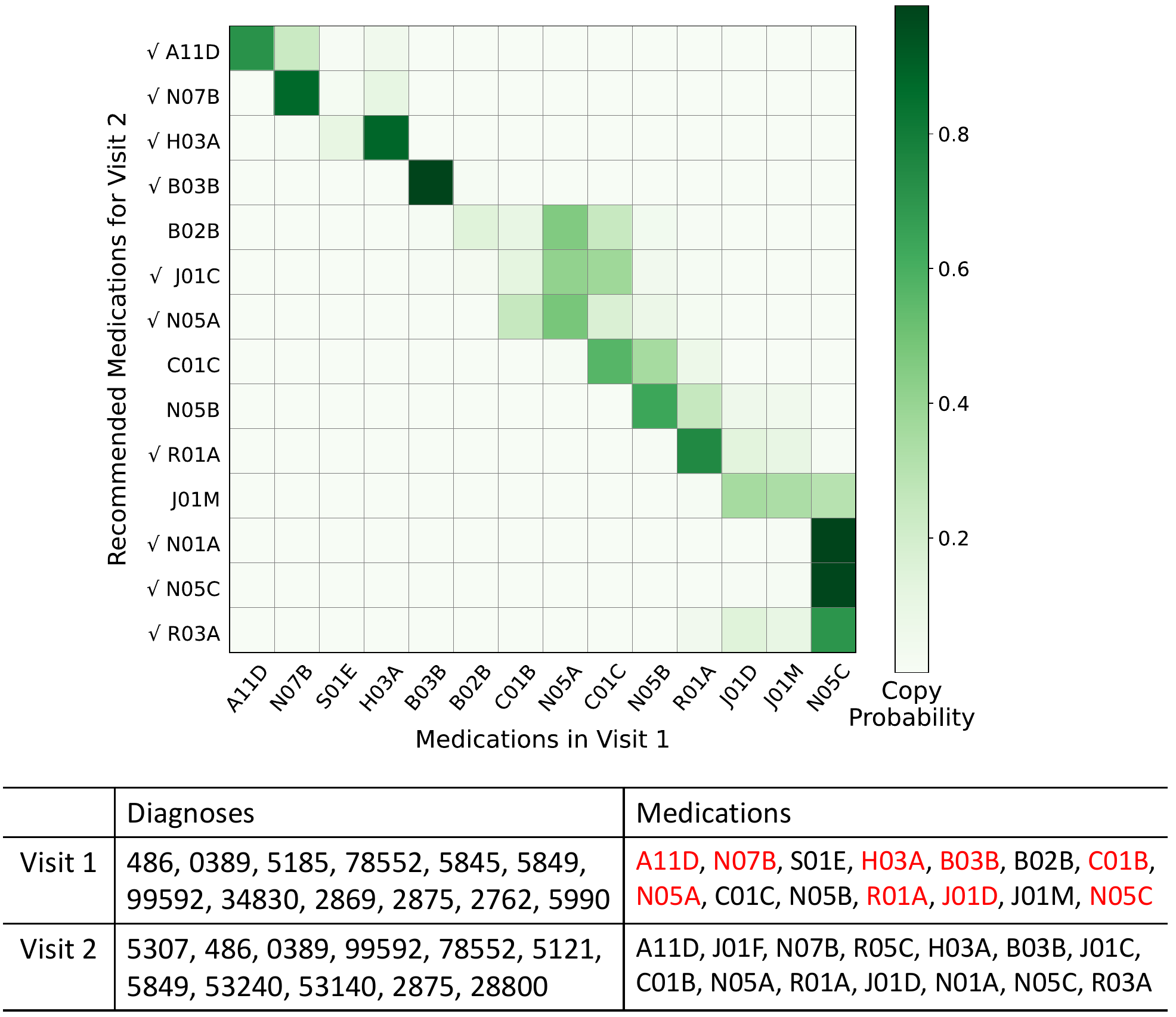}
    \caption{The example for case study. The table in the bottom shows the records of the patient and the red color indicates that the corresponding medication can be copied. The $\surd$ symbol means the corresponding recommendation is correct.}
    \label{fig:case study}
\end{figure}

\subsection{Case Study}

We present an example patient in MIMIC-III to illustrate how COGNet conducts the copy mechanism to improve medication recommendation. 
This patient visited the hospital twice. At the first time, the patient was mainly diagnosed with pneumonia, septicemia, sepsis, etc. Later, the patient returned for a second visit. In addition to the previously diagnosed diseases, the patient also had some new diseases, like gastroesophageal laceration and gastric ulcer. The table in Fig~\ref{fig:case study} shows the details. Due to space constraints, we use International Classification of Diseases (ICD)\footnote{\url{http://www.icd9data.com/}} codes to represent diagnosis results and Anatomical Therapeutic Chemical (ATC)\footnote{\url{https://www.whocc.no/atc/structure_and_principles/}} classification system to represent medications. 

As shown in Figure~\ref{fig:case study}, we visualize the copy probability $\Pr_c$ computed by Eq.(20) at each decoding step of recommending the medications for the second visit.
We can see that some reusable medications, like A11D, N07B and H03A, are correctly copied by assigning high probabilities to them in previous visit. 
In addition, some new drugs, like J01C and R03A, can also be appropriately generated. 
It indicates that COGNet can not only copy historical drugs according to unhealed diseases, but also generate new drugs based on new diagnosis results. 
The visualization results also hint that COGNet can provide a good way for the interpretation of the medication recommendation process.

\section{Related Work}
\subsection{Medication Recommendation}
Due the clinical value, medication recommendation has received increasing attention in recent years. According to the used information, existing approaches can be broadly categorized into rule-based, instance-based, and longitudinal methods. 
\subsubsection{Rule-based Methods} Rule-based methods \cite{almirall2012designing, ChenMSGT16, gunlicks2016pilot,LakkarajuR17} rely on the human-designed recommendation protocols. For example, Gunlicks-Stoessel et al. \cite{gunlicks2016pilot} attempts to recommend porper treatments for adolescent depression based on rules.
However, these methods require a lot of effort from clinicians and lack generalization. 
\subsubsection{Instance-based Methods} Instance-based methods~\cite{LEAP,gong2021smr} only take the information of current visit as input. For example, Zhang et al.~\cite{LEAP} firstly encode the patient's current diagnose and then use a recurrent decoder to generate the medication recommendations based on the encoded information. 
However, they ignore the historical visit information of the patient.
\subsubsection{Longitudinal Methods} These approaches \cite{WangZHZ18, WangRCR0R19, gamenet, changematter, safedrug, premier} use the historical information of patients and explore the sequential dependency between visits. 
Most of them basically model the longitudinal patient information by RNNs.
Le \textit{et al.} \cite{dmnc} and Shang \textit{et al.} \cite{gamenet} combine memory networks with RNNs to enhance the memory ability. 
Yang \textit{et al.} \cite{safedrug} further incorporate the drugs' molecule information to improve the medication representation learning.
Yang \textit{et al.} \cite{changematter} explicitly model the health condition changes of the patient to enhance the correlations between continuous visits by a recurrent residual learning approach.
However, these methods do not explicitly consider the relationship between the medication recommendations of the same patient and the RNN like methods are hard to handle the long range visit dependency.

\subsection{Graph Neural Network}
Recently, graph neural networks (GNN)~\cite{gcn,zhuang2018dual,hamilton2017inductive,atwood2016diffusion,Velickovic2017,DBLP:conf/recsys/VasileSC16} have received wide attention in many fields. 
The convolutional GNN can learn powerful node representations by aggregating the neighbors' features over the graph.
Some works~\cite{gamenet,safedrug} have attempted to leverage the GNN to improve the medication recommendation.
For example, Yang \textit{et al.} \cite{safedrug} take the drugs' molecule structures as graphs and use GNN to learn the better medication representations to improve medication recommendation.
Shang \textit{et al.}~\cite{gamenet} use the GNN to encode the EHR and DDI graphs to introduce the medication correlation information.
In this paper, inspired by ~\cite{gamenet}, we use the GNN to encode the medication co-occurrence and DDI relationships to improve the recommendation.

\section{Conclusion}
In this paper, we proposed a novel medication recommendation model, COGNet, to better leverage historical information from a medication-level perspective. COGNet works under an encoder-decoder based framework and introduces a \textit{copy-or-predict} mechanism to accurately determine whether a historical medication is still relevant at present. Experiment results on the publicly available MIMIC-III dataset demonstrate that COGNet outperforms existing medication recommendation methods. We also investigate the impact of number of visits on the performance, which shows that COGNet can effectively incorporate the information of multiple past visits.
Further ablation study results also suggest the effectiveness of each module of COGNet.


\begin{acks}
This work was supported by National Key R\&D Program of China, No. 2018YFC0117000.
This work was also partially supported by Natural Science Foundation of China grant, No. U21A20488.
\end{acks}

\bibliographystyle{ACM-Reference-Format}
\bibliography{main}


\appendix
\section{Additional Experimental Setups}
\subsection{Dataset Processing}
In this section, we elaborate the operation of data processing. The original data is from “ADMISSIONS.csv”, “DIAGNOSES\_ICD.csv“, “PROCEDURES\_ICD.csv“ and ”PRESCRIPTIONS.csv“ files from the 1.40 version MIMIC-III. These tables are merged through admission id and subject id ("Admission" has the same meaning as "Visit" in this paper). We utilized all the patients with at least 2 visits. We prepared the available medications by retaining the top 300 medications in terms of number of occurrences. This is to improve the training speed for easier analysis. We extracted Top-40 severity DDI types from TWOSIDES \cite{ddi}, which are reported by ATC Third Level codes. In order to be able to compute DDI score, we transform the NDC drug codes to same ATC level codes. After the above operation was completed, we divided the data into training, validation and test by the ratio of $\frac{2}{3}:\frac{1}{3}:\frac{1}{3}$. We counted the frequency of all drug occurrences on the training set,and then resort the medications in ascending order by frequency for all patients.

\subsection{Implementation Details}
Our method is implemented by PyTorch 1.9.0 based on python 3.9.6, tested on an Intel Xeon 8255C machine with 315G RAM and 8 NVIDIA Tesla V100 GPUs. We choose the optimal hyperparameters based on the validation set, where the dimension size $s=64$, number of beam search states is 4 and maximum generation length is 45. Models are trained on Adam optimizer \cite{adam} with learning rate $1\times10^{-4}$ and batch size 16 for 50 epochs. We fixed the random seed as 1203 for PyTorch to ensure the reproducibility of the models.
During the test process, for a fair comparison, we apply bootstrapping sampling instead of cross-validation according to \cite{safedrug}. Precisely, we random sample $80\%$ data from test set for a round of evaluation and the results of 10 rounds are used to calculate the mean and standard deviation, which are finally reported. As the sampling process is random, the final experimental results may vary slightly.

\subsection{Metrics}
In this section, we present the definition of each metric used in the experiment section.
\begin{itemize}
    \item \textbf{Jaccard} for a patient is calculated as below:
    \begin{equation}
        \operatorname{Jaccard} = \frac{1}{T} \sum_{i=1}^T \frac{|\mathcal{M}_i \cap \hat{\mathcal{M}}_i|}{|\mathcal{M}_i \cup \hat{\mathcal{M}}_i|}
    \end{equation}
    where $\mathcal{M}_i$ is the ground-truth medication combination and $\hat{\mathcal{M}}_i$ is the predicted result.
    
    \item \textbf{F1} is the harmonic mean of precision and recall. For a patient, it is calculated as follows:
    \begin{equation}
        \operatorname{Precision_i} = \frac{|\mathcal{M}_i \cap \hat{\mathcal{M}}_i|}{|\hat{\mathcal{M}}_i|}
    \end{equation}
    \begin{equation}
        \operatorname{Recall_i} = \frac{|\mathcal{M}_i \cap \hat{\mathcal{M}}_i|}{|\mathcal{M}_i|}
    \end{equation}
    \begin{equation}
        \operatorname{F1} = \frac{1}{T} \sum_{i=1}^T  2*\frac{\operatorname{Precision_i}*\operatorname{Recall_i}}{\operatorname{Precision_i} + \operatorname{Recall_i}}
    \end{equation}
    
    \item \textbf{PRAUC} refers to Precision Recall Area Under Curve. To compute PRAUC, each medication should correspond to a probability to be recommended \cite{safedrug}. However, since we model the medication recommendation as a sequential generation problem, it means that each medication will have a probability at every generation step. To resolve this issue, we adopt the following methods. For each recommended medications, we directly take the probability corresponding to the step in which they are recommended. For the other medications, we take the average of probabilities at all steps. Then the PRAUC can be calculated by
    \begin{equation}
        \begin{split}
        \operatorname{PRAUC} = \frac{1}{T} \sum_{i=1}^T \sum_{k=1}^{|\mathcal{M}|}\operatorname{Precision}(k)_i \Delta\operatorname{Recall}(k)_i \\
        \Delta\operatorname{Recall}(k)_i = \operatorname{Recall}(k)_i - \operatorname{Recall}(k-1)_i
        \end{split}
    \end{equation}
    where $k$ is the rank in the sequence of drugs. $\operatorname{Precision}(k)_i$ is the precision at cut-off $k$ in ordered retrieval list, and $\Delta\operatorname{Recall}(k)_i$ is the change in recall when deriving the $k$-th drug.
    
    \item \textbf{DDI} measure the interaction between the recommended medications, which is calculated by:
    \begin{equation}
        \operatorname{DDI} = \frac{1}{T} \sum_{i=1}^T \frac{\sum_{j=1}^{|\hat{\mathcal{M}}_i|} \sum_{k=j+1}^{|\hat{\mathcal{M}}_i|} \textbf{1}\{\mathbf{A}_d[\hat{\mathcal{M}}_i^{(j)},\hat{\mathcal{M}}_i^{(k)}]=1\}}{\sum_{j=1}^{|\hat{\mathcal{M}}_i|} \sum_{k=j+1}^{|\hat{\mathcal{M}}_k|} 1}
    \end{equation}
    where $\mathbf{A}_d$ is the adjacency matrix of DDI graph defined in section 4, $\hat{\mathcal{M}}_i^{(j)}$ denoted the $j$-th recommended medication and $\textbf{1}\{\cdot\}$ is a function that return 1 when expression in $\{\cdot\}$ is true, otherwise 0.
\end{itemize}

\section{Additional Experiments}
\subsection{Performance with Different Label Order}

\begin{figure}
    \centering
    \includegraphics[width=\linewidth]{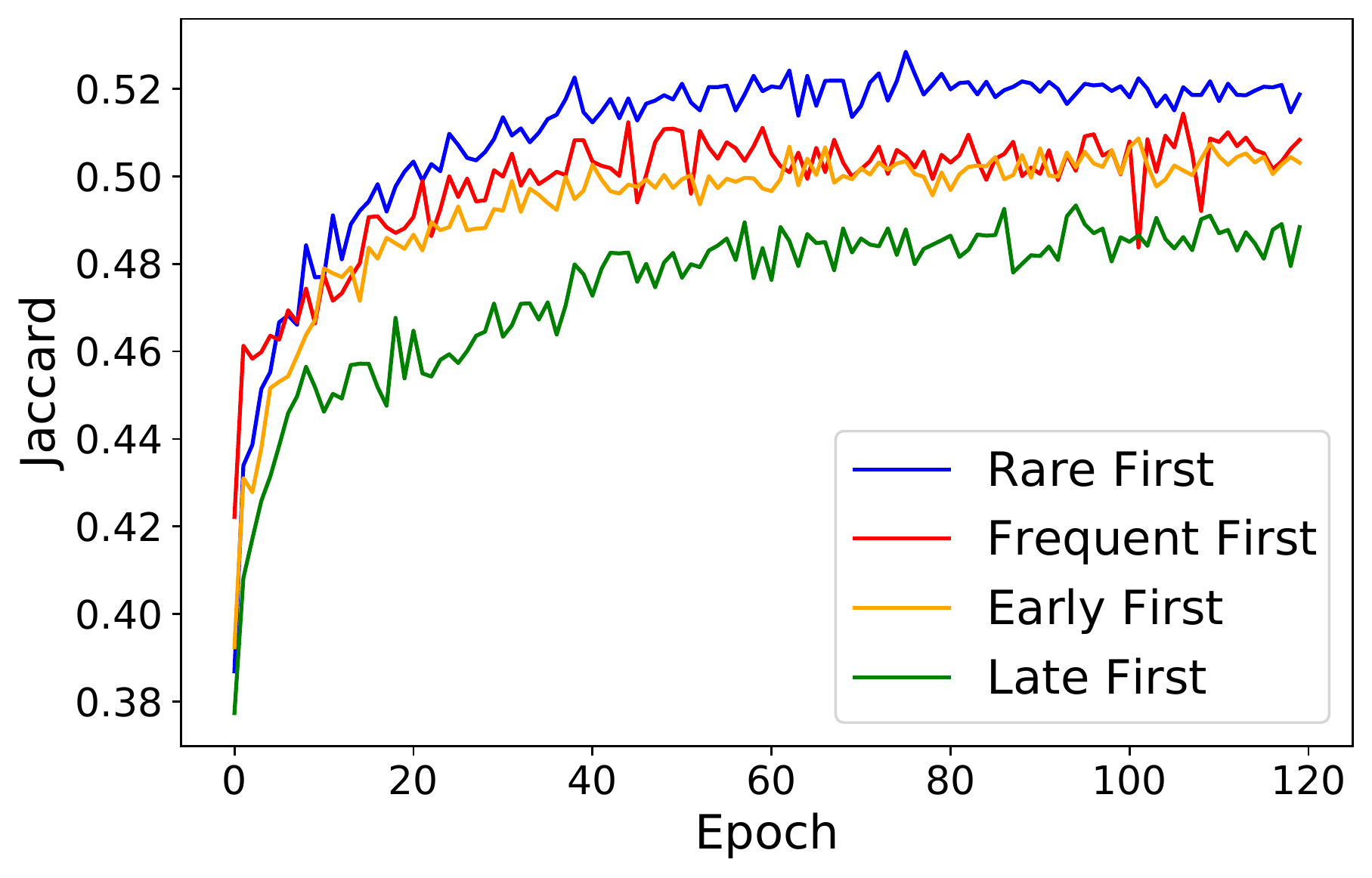}
    \caption{Performance w.r.t training epochs on MIMIC-III with different label order.}
    \label{fig:order}
\end{figure}

Since new medications are recommended under the conditions of already recommended medications, the relative order of the medications may have an impact on the final results. To explore which kind of sequencing would be more appropriate of medication recommendation, we conduct four different heuristics to order the data:
\begin{itemize}
    \item \textbf{Rare first} ranks medications by their frequencies in the training data, placing the those that appear less frequently first.
    \item \textbf{Frequent first} put frequent medications before more rare medications.
    \item \textbf{Early first} sorts medications by chronological order, with those used earlier placed first.
    \item \textbf{Late first } arranges the medications in reverse chronological order.
\end{itemize}

Fig~\ref{fig:order} shows the performance of COGNet with different ordering heuristics over training epochs. \textit{Rare first} outperform the alternative labeling strategies, because it allows the model to focus more on unusual medications, thereby alleviating the data imbalance problem. \textit{Frequent first} converges faster but performs poorly. The main reason is that the recommended medications are predominated by non-informative popular medications. The final results of \textit{Early first} and \textit{Late first} are weak, which indicates that chronological order is not a good choice in this task. In addition, we tried to randomly disrupt all the drugs, which is far worse than the above heuristics, so we do not show it. 
\end{document}